\documentclass[conference]{IEEEtran}
\IEEEoverridecommandlockouts
\usepackage{cite}
\usepackage{amsmath,amssymb,amsfonts}
\usepackage{algorithmic}
\usepackage{graphicx}
\usepackage{textcomp}
\usepackage{xcolor}
\def\BibTeX{{\rm B\kern-.05em{\sc i\kern-.025em b}\kern-.08em
    T\kern-.1667em\lower.7ex\hbox{E}\kern-.125emX}}

\makeatletter
\newcommand{\linebreakand}{%
  \end{@IEEEauthorhalign}
  \hfill\mbox{}\par
  \mbox{}\hfill\begin{@IEEEauthorhalign}
}
\makeatother

\begin{document}

\title{
Simple Physical Adversarial Examples against End-to-End Autonomous Driving Models}

\author{\IEEEauthorblockN{Adith Boloor}
\IEEEauthorblockA{\textit{Electrical and Systems Engineering}\\
\textit{Washington University in St. Louis}\\
St. Louis, USA \\
a.jagadi@wustl.edu}

\and

\IEEEauthorblockN{Xin He}
\IEEEauthorblockA{\textit{Computer Science and Engineering} \\
\textit{University of Michigan, Ann Arbor}\\
Ann Arbor, USA \\
xinhe@umich.edu}

\and

\IEEEauthorblockN{Christopher Gill}
\IEEEauthorblockA{\textit{Computer Science and Engineering} \\
\textit{Washington University in St. Louis}\\
St. Louis, USA \\
cdgill@wustl.edu}

\linebreakand 

\IEEEauthorblockN{Yevgeniy Vorobeychik}
\IEEEauthorblockA{\textit{Computer Science and Engineering} \\
\textit{Washington University in St. Louis}\\
St. Louis, USA \\
yvorobeychik@wustl.edu}

\and

\IEEEauthorblockN{Xuan Zhang}
\IEEEauthorblockA{\textit{Electrical and Systems Engineering} \\
\textit{Washington University in St. Louis}\\
St. Louis, USA \\
xuan.zhang@wustl.edu}
}

\maketitle

\begin{abstract}
Recent advances in machine learning, especially techniques such as deep neural networks, are promoting a range of high-stakes applications, including autonomous driving, which often relies on deep learning for perception.
While deep learning for perception has been shown to be vulnerable to a host of subtle adversarial manipulations of images, end-to-end demonstrations of successful attacks, which manipulate the physical environment and result in physical consequences, are scarce.
Moreover, attacks typically involve carefully constructed adversarial examples at the level of pixels.
We demonstrate the first end-to-end attacks on autonomous driving in simulation, using simple physically realizable attacks: the painting of black lines on the road.
These attacks target deep neural network models for end-to-end autonomous driving control.
A systematic investigation shows that such attacks are surprisingly easy to engineer, and we describe scenarios (e.g., right turns) in which they are highly effective, and others that are less vulnerable (e.g., driving straight).
Further, we use network deconvolution to demonstrate that the attacks succeed by inducing activation patterns similar to entirely different scenarios used in training.
\end{abstract}


\section{Introduction}
With billions of dollars being pumped into autonomous vehicle research to reach Level 5 Autonomy, where vehicles will not require human intervention, safety has become a critical issue.
The remarkable advances in deep learning, in turn, suggest such approaches as natural candidates for integration into autonomous control.
One way to use deep learning in autonomous driving control is in an end-to-end (e2e) fashion, where learned models directly translate perceptual inputs into control decisions, such as steering angle.
Indeed, recent work demonstrated such approaches to be remarkably successful, particularly when learned to imitate human drivers~\cite{Bojarski2017ExplainingHA}.

\begin{figure}
  \centering  \includegraphics[width=0.9\columnwidth]{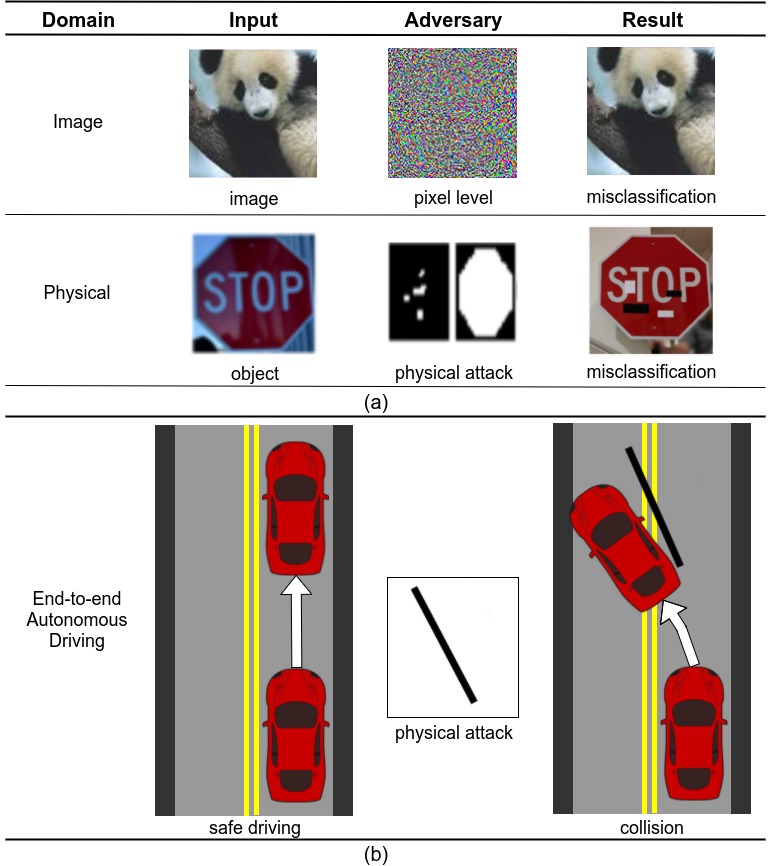}
  \caption{(a) Existing attacks on machine learning models in the image ~\cite{Goodfellow2015ExplainingAH} and the physical domain ~\cite{Eykholt2017RobustPA}; (b) conceptual illustration of potential physical attacks in the end-to-end driving domain studied in our work.}
  \label{fig:conceptual_overview}
\end{figure}

Despite the success of deep learning in enabling greater autonomy, a number of parallel efforts also exhibited concerning fragility of these approaches to small adversarial perturbations of inputs, such as images~\cite{Vorobeychik18book,Dreossi2018SemanticAD}.
Moreover, such perturbations have been shown to effectively translate to physically realizable attacks on deep models, such as by putting stickers on stop signs to cause miscategorization of these as speed limit signs~\cite{Eykholt2017RobustPA}.
Fig.~\ref{fig:conceptual_overview}(a) offers several canonical illustrations.

There is, however, a crucial missing link in most adversarial example attacks to date:  manipulations of the physical environment that have a demonstrable \emph{physical} impact (e.g., a crash).
For example, typical attacks consider only prediction error as an outcome measure and focus either on a static image, or a fixed set of views, without consideration of the dynamics of closed-loop autonomous control.
To bridge this gap, our aim is to study  \emph{end-to-end} adversarial examples.
We require such adversarial examples to satisfy four criteria: 1) modify physical environment, 2) be simple to implement, 3) appear unsuspicious, and 4) have a physical impact, such as causing a lane violation or a crash.
The prevalent attacks that introduce carefully engineered manipulations fail the simplicity criterion~\cite{Papernot2016TheLO,Vorobeychik18book}, whereas the simpler physical attacks, such as stickers on a stop sign, are evaluated solely on prediction accuracy~\cite{Eykholt2017RobustPA}.

The particular class of attacks we systematically study is the painting of black lines on the road, as shown in Fig.~\ref{fig:conceptual_overview}(b).
These are unsuspicious since they are semantically inconsequential (few human drivers would be confused) and similar to common imperfections seen in the wild, such as tread marks.
Furthermore, we demonstrate a systematic approach for designing such attacks so as to maximize steering angle, and demonstrate actual physical impact (lane violations and crashes) over a variety of scenarios, in the context of state-of-the-art end-to-end deep learning-based controllers.
We consider scenarios where correct behavior involves turning right, left, and going straight.
Surprisingly, we find that right turns are by far the riskiest; on the other hand, as expected, going straight is rather robust to our class of attacks.

Our final contribution is to use network deconvolution to explore the reasons behind successful attacks.
Here, our findings suggest that one of the causes of controller failure is in partially mistaking painted lines on the road for a barrier common during left-turn scenarios, thereby causing the car to steer sharply left when it would otherwise turn right.

\section{Background and Related Work}

\subsection{Deep Neural Networks for Perception and Control}
Neural network (NN) algorithms are loosely modeled after the human brain which allows them to recognize patterns in high-dimensional data. 
To address large complex problems, Deep Neural Networks (DNNs) are designed with a deeper and wider hierarchy so that the network model has a larger learning capability to accommodate diverse inputs with more features. 
They have been used to achieve a high level of accuracy in perception related tasks such as image classification \cite{Krizhevsky2012ImageNetCW} and semantic segmentation \cite{Krizhevsky2012ImageNetCW}.


End-to-end (e2e) learning models comprises DNNs that accept raw input parameters in one end and directly calculate the desired output at the other end. Rather than explicitly decomposing a complex problem into its constituent parts and solving them separately, e2e models directly generate the output from the inputs. It is achieved by applying gradient-based learning to the system as a whole. Recently \cite{Bojarski2016EndTE}, e2e models have been shown to have good performance in the domain of autonomous vehicles, where the forward facing camera input can be directly translated to control (steer, throttle and brake) commands. 

\subsection{Attacks on Deep Learning for Perception and Control}
Attacks or adversarial examples \cite{Vorobeychik18book, Lowd2005AdversarialL} are deliberately calculated perturbations to the input which result in an error in the output from a trained DNN model. 

The idea of using adversarial examples against static image classification models has been studied and it has been proved that DNNs are highly susceptible to carefully designed pixel-level adversarial perturbations\cite{Papernot2016TheLO, Goodfellow2014GenerativeAN}. 
Perturbed images that would be easily ignored by humans may not be correctly recognized by the DNN model. 
More recently, attacks in the physical domain have begun to draw more attention. Studies show that adding stickers to a stop sign in carefully positioned ways can result in the classification model to mis-identify the stop sign to be a speed-limit sign.
However, existing investigations on adversarial examples still focus on classification errors associated with static images and are conducted in limited experimental environments ~\cite{Eykholt2017RobustPA, Lu2017NONT, Dreossi2018SemanticAD}.
Research considering the learning model in a dynamic system setting, like on autonomous vehicles in the real world is sparse \cite{Tuncali2018SimulationbasedAT}. 
In this paper, we aim to address these current limitations and provide a methodology to systematically study physically realizable attacks on the e2e models in realistic driving conditions.


\section{Modeling Framework}

\begin{figure*}
  \centering  \includegraphics[width=\textwidth]{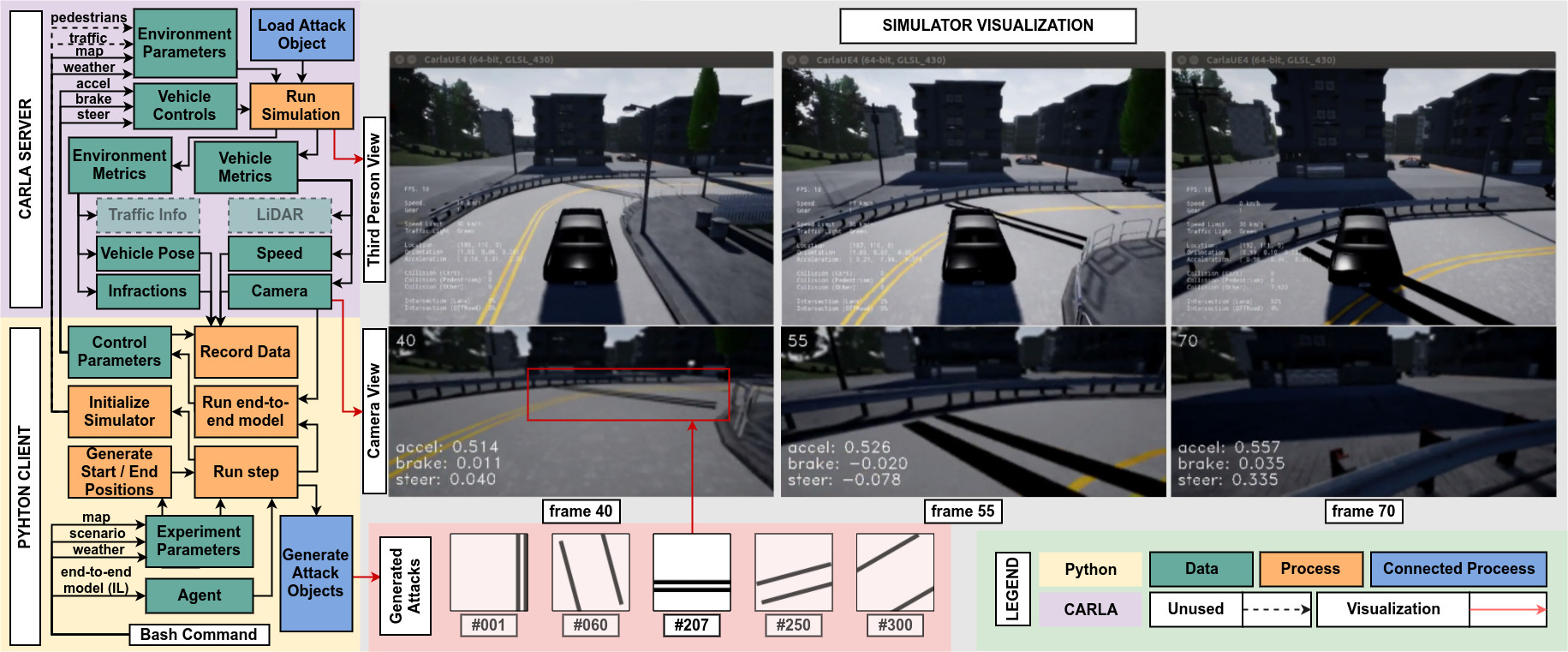}
  \caption{Architecture overview of our simulation infrastructure including the interfaces between the CARLA simulator and the pattern generator scripts. Visualization of the camera and the third person views from one attack episode are also shown.}
  \label{fig:overall_architecture}
\end{figure*}

In this paper, we focus on exploring the influence of a physical adversary that successfully subverts RGB camera-based e2e driving models. We define physical adversarial examples as attacks which are physically realizable in the real world. For example, deliberately painted shapes on the road or on stop signs, would be classified as physical adversaries. Fig. \ref{fig:conceptual_overview}(b) displays the conceptual view of such an attack involving painting black lines. We define prospective adversarial examples as \textit{patterns}. To create an adversarial example that forces the e2e model to crash the vehicle, we need to choose the parameters of \textit{pattern}'s shape that maximize the difference between the steering values between the ground truth (case without an attack) and the one with the attack.
For a particular task where the e2e model has to drive the vehicle forward while maintaining its own lane, we need to choose an attack that causes the steering angle value to increase (or decrease) continuously. This would cause the vehicle to veer into the wrong lane or go offroad, which we characterize as a successful attack. Conventional gradient descent based attack techniques cannot be applied in this domain since the generated attacks requires pixel-level modifications spanning the entire input space, which is not physically realizable.
Therefore, our solution is to systematically explore a more confined search space. We choose regions of interest on the road where we will create attacks, and we begin by \textit{drawing} a simple pattern like a thin black strip with fixed width and length on those regions. Then we sweep through different positions and orientations of the pattern to see if exhaustively going through the search space finds adversarial examples that cause the vehicle to crash. Note that for the entirety of this research, we use the right-driving traffic system.

At the high level, our goal is to paint a pattern (such as the black strip) somewhere on the track (road) to cause a crash. We formalize the latter objective as that of maximizing or minimizing the induced steering angle, since, assuming acceleration remains the same with an attack as without it, this is sure to cause the car to veer off the road.
Since the problem is dynamic, we must consider the impact of the object we paint on the track over a sequence of frames that capture the track, along with this pattern, as the vehicle moves towards and, eventually, over the modified track segment.
Crucially, we modify the track itself, which is subsequently captured in vision, digitized, and used as input into the e2e model's controller.

To formalize, we now introduce some notation.
Let $l$ denote the position on the track where we place the pattern, which we, in turn, denote by $\delta$.
We use $L$ to denote the set of feasible locations at which we can position the adversarial pattern $\delta$, and $S$ the set of possible patterns (along with associated modifications; in our case below, for example, we consider either a single black line, or a pair of parallel black lines, with modifications involving, for example, length of the line and its rotation angle).
Let $x_l$ be the state of the track at position $l$, and $x_l +\delta$ then becomes the state of the track at this same position when the pattern $\delta$ is added to it.
The state of the track at position $l$ is captured by the vehicle's vision system when it comes into view; we denote the frame at which this location initially comes into view by $f_l$, and let $\Delta$ be the number of frames over which the track in position $l$ is visible to the vehicle's vision system.
Given the track state $x_l$ at position $l$, the digital view of it in frame $f$ is denoted by $y_f(x_l)$.
Finally, we let $f_{sa}(y_f,h_f)$ denote the predicted steering angle given observed digital image corresponding to frame $f$, and prior history of frames, $h_f$.
We can formulate the optimization problem we aim to solve as follows:
\begin{subequations}
\begin{align}
\label{E:attack}
\mathrm{Collide\ Right:\ }&\max_{l,\delta} \sum_{\tau = 0}^{\Delta} f_{sa}(y_{f_l + \tau}(x_l+\delta), h_{f_l + \tau})\\
\mathrm{Collide\ Left:\ }&\min_{l,\delta} \sum_{\tau = 0}^{\Delta} f_{sa}(y_{f_l + \tau}(x_l+\delta), h_{f_l + \tau})\\
&\mathrm{subject\ to:}\quad l \in L, \quad \delta \in S.
\end{align}
\end{subequations}

Essentially, equation （\ref{E:attack}） says that to optimize an attack that causes the vehicle to veer off towards the right and collide, we need to maximize the sum of steering angles for that particular experiment for the frames in which the pattern is in view. And similarly, we need to minimize the steering sum, to make the vehicle veer left.
The validity of this optimization objective is evaluated in Section~\ref{sec:res}.

\section{Experimental Methodology}

This section introduces the various building blocks that we used to perform our experiments. Fig.~\ref{fig:overall_architecture} shows the overall architecture of our experimentation method, including the CARLA simulator block, the python client block, and how they communicate with each other to test the patterns on the simulator.

\subsection{Autonomous Vehicle Simulator}
Simulators have been used to test autonomous vehicles to for the sake of efficiency and safety \cite{Shah2017AirSimHV, Fan2018BaiduAE, Tian2018DeepTestAT}. We ran our experiments on the CARLA \cite{Dosovitskiy2017CARLAAO} autonomous vehicle simulator. Built using Unreal Engine 4 \cite{UnrealEngine4}, CARLA has sufficient flexibility to create reasonably realistic simulated environments, with a robust physics engine, lifelike lighting, 3D objects including roads, buildings, traffic signs, and non-player characters including pedestrians and other vehicles. Fig.~\ref{fig:overall_architecture} shows how the simulator looks in the third person view. It allows us to acquire sensor data like the camera image for each frame (camera view), vehicle measurements (speed, steering angle and brake) and other environmental metrics like how the vehicle interacts with the environment in the form of infractions and collision intensity. Since we are using e2e models that are use only the RGB camera, we disabled the LiDAR, semantic segmentation, and depth cameras. Steering angle, throttle and brake parameters are the primary control parameters for driving the vehicle in the simulation. CARLA (stable version 0.8.2 as of writing this paper) comes with two fully built maps: a large training map and a smaller testing map which were used for training and testing the e2e models respectively. CARLA also allows the user to run experiments under various weather conditions like sunset, cloudy and rain, which are determined by the client input. To keep a consistent frame rate and execution time, we run CARLA using a fixed time-step.

\subsection{End-to-end Driving Models}
The CARLA simulator comes with two trained end-to-end models: Conditional Imitation Learning (IL) \cite{Codevilla2018EndtoEndDV} and Reinforcement Learning (RL) \cite{Dosovitskiy2017CARLAAO}. Their commonality ends at using the camera image as the input and producing output controls that include steering angle, acceleration, and brake. The IL model uses a trained model consisting of demonstrations of human driving on the simulator. In other words, the IL model tries to mimic the actions of the expert with whom it was trained with. RL uses a trained deep network based on a rewards system, provided by the environment based on the corresponding actions, without the aid of human drivers. More specifically, for RL, the asynchronous advantage actor-critic (A3C) algorithm was used. It is worth mentioning that IL performed better than RL in untrained scenarios \cite{Dosovitskiy2017CARLAAO}.

\subsection{Physical Adversary Generation}
To generate physically realizable adversaries in a systematic manner, we first modify the original CARLA maps so that we can place the aforementioned \textit{patterns} wherever we need. We build a pattern generator that can create different kinds of shapes (single and double lines with various attributes) using the pattern parameters. We create a 200 x 200 pixel region on the road which matches the width of the road. This canvas is mapped to the pattern file read from the server, and is placed in the simulation. For the pattern generator, we explore parameters like the position, width, and rotation of the line. To generate different variations on the attack, we swept the pattern from one side of the road to the other (position 0 to 200), and varied the rotation between 0 and 180 degrees for each step. Similarly, we created a more advanced pattern which involves two parallel black lines we call the \textit{double-line} pattern. It comprises the previous parameters, viz., position, rotation, and width, with the addition of the new gap parameter which is the distance between the two parallel lines. Fig.~\ref{fig:overall_architecture} shows some examples of the generated double line patterns which can be seen overlaid on the road in frames 55 and 70.	

\subsection{Data Collection and Processing}
To ensure a broad scope to test the effectiveness of the different attacks in various settings, we conduct experiments by changing various environment parameters like the maps (training map and testing map), scenes, weathers (clear sky, rain, and sunset), driving scenarios (straight road, right corner and left corner), e2e models (IL and RL) and the entire search space for the patterns. To be able to search the design space thoroughly, we prepare a CARLA docker which allow us to run as many as 16 CARLA instances simultaneously, spread out over 8 RTX GPUs\cite{RTX}. 
We choose the baseline scenarios (no attack) where the e2e models drive the vehicle with minimal infractions. We ran the experiments at 10 fps, and collected the following data for each camera frame (a typical experiment ran between 60 to 100 frames): camera image from the mounted RGB camera, vehicle speed, predicted acceleration, predicted steering, predicted braking, percentage of vehicle on the wrong lane, percentage of the vehicle on the sidewalk (offroad), and collision intensity. Fig.\ref{fig:overall_architecture} also shows this dataflow which suffices to assess the ramifications of a particular pattern in a certain experiment. 

\section{Experimental Results}
\label{sec:res}
Through experimentation, we demonstrate the existence of conspicuous physical adversaries that successfully break the e2e driving models. These adversaries do not need to be subtle or sophisticated modifications. Although they can be easily distinguished and thus ignored by humans drivers, they are effective in generating serious traffic infractions for the e2e autonomous driving models we have evaluated.

\subsection{Summary of Physical Adversaries}

\begin{figure}
  \centering  \includegraphics[width=\columnwidth]{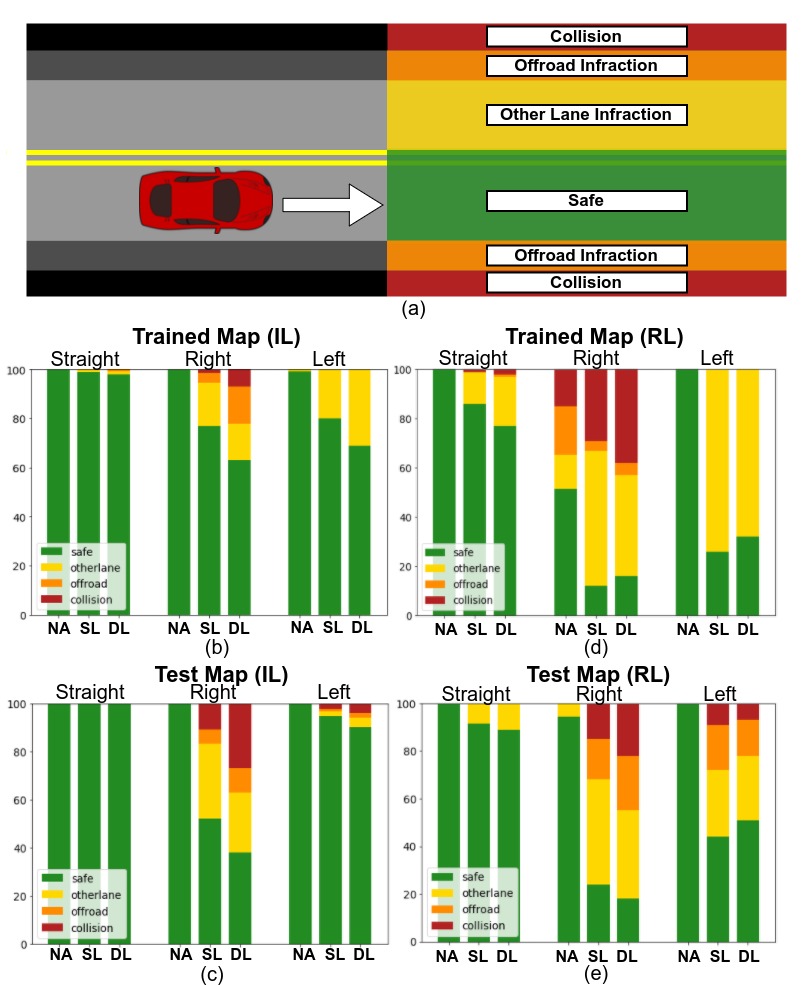}
  \caption{Comparison of the infractions caused by different patterns. (a) Driving Infraction regions; (b)(c) Infraction percentages for IL; (d)(e) Infraction percentages for RL; NA - No Attack, SL - Single Line pattern, DL - Double Lines pattern; Straight - Straight Road Driving, Right - Right Corner Driving, Left - Left Corner Driving}
  \label{fig:result01}
\end{figure}

We generate two primary sets of adversary patterns: single line with varying positions and rotation angles, and double lines with varying positions, rotation angles and distance (gap) between the lines. In Fig.~\ref{fig:result01}(a), we define different safety regions of the road in ascending order of risk. We start with the vehicle's own lane (safe region), the opposite lane (unsafe), offroad/sidewalk (dangerous) and regions of collisions (very dangerous) past the offroad region. Fig.\ref{fig:result01}(b)(c)(d)(e) shows that by sweeping through the three scenarios (straight road driving, right corner driving, left corner driving) with the single and double line patterns, for both the training map and testing maps, we see that some patterns cause infractions. 
First, we observe the transferability of adversaries since some of our generated adversarial examples cause both IL (Fig.\ref{fig:result01}(b)) and RL (Fig.\ref{fig:result01}(d)) models to produce infractions. Second, the IL model performs better than its RL counterpart. Additionally, we notice that the double line adversarial examples cause more severe infractions than its single line counterpart. Lastly, we observe that \textit{Straight Road Driving} and \textit{Left Corner Driving} are more resilient to level 2, and level 3 infractions, hence, in the next section, we analyze the cases for the \textit{Right Corner Driving} case with Imitation Learning more thoroughly.

\subsection{Analysis of Attack Objectives}
\begin{figure}
  \centering  \includegraphics[width=\columnwidth]{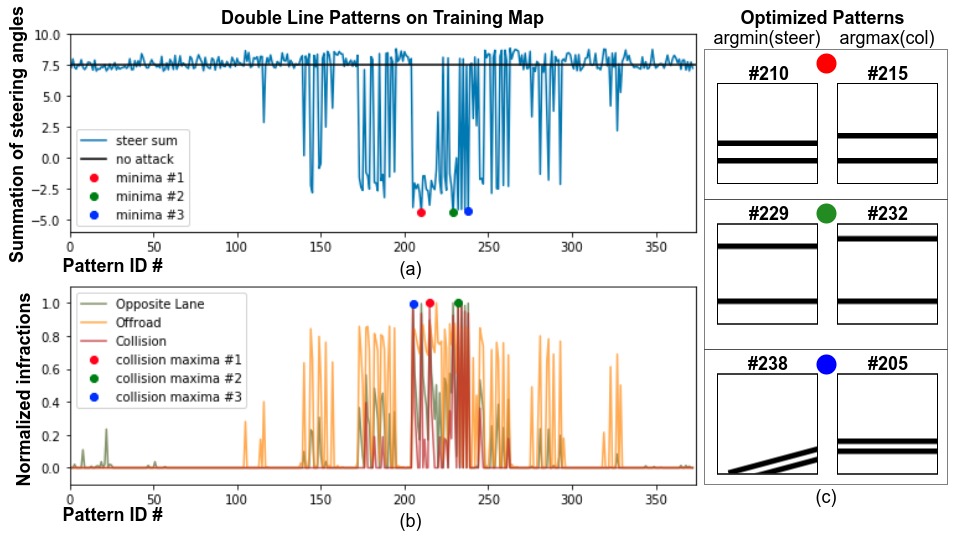}
  \caption{Adversary against 'Right Corner Driving'. (a) Adversarial examples significantly changes the steering control. (b) Some patterns cause minor infractions whereas others cause level 3 infractions. (c) The patterns that cause the minimum steering sum and maximum collisions look similar.}
  \label{fig:result02}
\end{figure}

To find the optimal adversary which would produce a level 3 infraction, i.e., a collision, for the scenario involving \textit{Right Turn Driving}, we have to find a pattern which would minimize the sum of steering angles as hypothesized in equation (\ref{E:attack}). A positive steering angle denotes steering towards the right and a negative steering angle implies steering towards the left. Fig.~\ref{fig:result02}(a)(b) show the steering sum and infraction sums respectively, over the course of 375 combinations of double line patterns. The infractions are normalized because collision data is recorded in SI units of intensity [kg*m/s], whereas the lane infractions are in percentages of the vehicle area in the respective regions. It also shows the three lowest points (minima) for the steering plot and the three highest points (maxima) for the collisions plot. In Fig.~\ref{fig:result02}(c), we use the \textit{argmin} and \textit{argmax} to observe the shapes of these adversarial examples. We observe that the \textit{patterns that minimize the sum of steering angle and correspondingly maximize the collision intensity are very similar.} 

\begin{figure}
  \centering  \includegraphics[width=0.9\columnwidth]{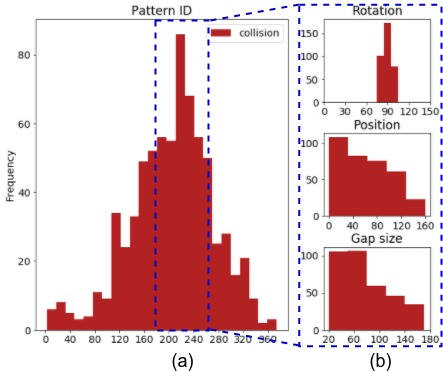}
  \caption{(a) Histogram showing strong adversaries. (b) Depiction of range of rotation, position and gap parameters for the most robust adversaries.}
  \label{fig:result2b}
\end{figure}

After gaining an intuition that some patterns perform better than others, we quantitatively analyze the range of parameters including rotation angles, position and gap size that will generate the most robust attacks, i.e., attacks that would perform well against different environmental conditions for the same scenario of right corner driving. Fig.~\ref{fig:result2b} shows a histogram of the collision amount versus the pattern IDs, and its corresponding parameters. We detect peaks in the histogram (Fig.~\ref{fig:result2b}(a)) which points out the fact that some patterns cause more infractions than others. Fig.~\ref{fig:result2b}(b) shows that some parameters play a stronger role than the others when it comes to generating an adversarial example. For example, pattern IDs between 180 and 260 are the most robust adversaries. These adversaries have a narrow range of rotation angles (90 - 115 degrees). Fig.~\ref{fig:result2b}(b) also shows that smaller gap sizes perform slightly better than larger ones.  

To get a stronger, underlying understanding of why these attacks work in the first place, and why some of them work better than the others, we peel through the layers of the e2e imitation learning network.

\subsection{Interpreting Attacks with DeConvNet} 
\begin{figure*}
  \centering  
  \includegraphics[width=\textwidth]{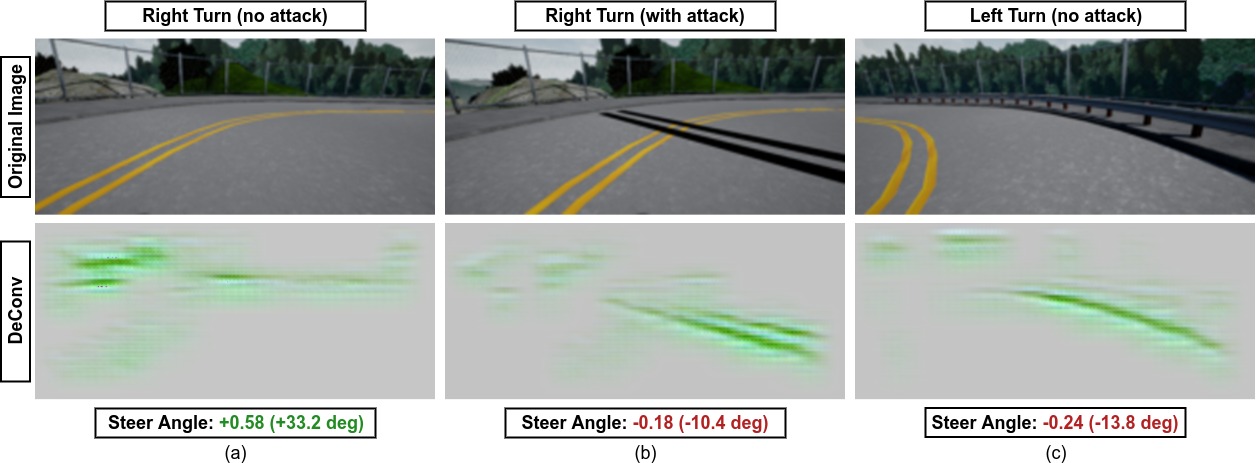}
  \caption{Attacks against Right Corner Driving: The top row shows the camera input while the bottom deconvolution images show that the reconstructed inputs from the strongest activations determine the steering angle. (a) Right Corner Driving without attack, (b) Right Corner Driving with attack and (c) Left Corner Driving without attack for comparison}
  \label{fig:result03}
\end{figure*}

To better understand the working mechanisms of the successful attack to the underlying imitation learning algorithm, the activities of feature maps inside the network need to be interpreted. Interpreting the activations requires mapping the feature maps to the input layer, hence we adopt a state-of-the-art technique, DeConvNet \cite{Zeiler2014VisualizingAU} to perform the mapping. We attach each CONV block to a DeConv counterpart, since the backbone of the imitation learning algorithm is a convolutional neural network which consists of eight CONV blocks for feature extraction and two fully connected (FC) blocks for classification. Each DeConv block uses the same filters, batchnorm parameters and activation functions as the CONV block except the operations are reversed. In this paper, DeConvNet is used merely as a probe to the already trained imitation learning network: it provides a continuous path to map high-level feature maps down to the input image. To interpret the network, the imitation learning network first processes the input image and computes the feature maps throughout the network layers. To view selected activations in the feature maps of a layer, other activations are set to zero, and the feature maps  backtrack through the rectification, reverse-batchnorm and transpose layers. Then, activations that contribute to the chosen activations in the lower layer are reconstructed. The process is repeated until the input pixel space is reached. Finally, the input pixels which give rise to the activations are visualized. In this experiment, we chose the \textit{top-200} strongest/largest activations in the \textit{fifth} convolution layer and mapped these activations down to the input pixel space for visualization. The reasons behind this choice are twofold: 1) The strongest activations stand out and dominate the decision-making in NNs and the \textit{top-200} activations are sufficient to cover the important activations. 2) Activations of the fifth CONV layer are more representative than other layers, since going deeper would mean that the amount of non-zero activations reduces significantly which invalidates the deconvolution operations, while shallow layers fail to fully capture the relation between different extracted features. 

We conduct a case study to understand why an attack works. Specifically, we take a deeper look inside the imitation network when adversaries are attacking the autonomous driving model for the scenario: right corner driving. The baseline case without any attack is depicted in Fig.~\ref{fig:result03}(a) while the one with double-line attack is shown in Fig.~\ref{fig:result03}(b). In the first row of Fig.\ref{fig:result03}, the input images from the front camera mounted on the vehicle are displayed, which are fed to the imitation learning network. In Fig. \ref{fig:result03}(a), the imitation learning network guides the vehicle to turn right at the corner, as the steering angle output is set to a positive value (steering +0.58). The highlighted green regions in the reconstructed inputs in the corresponding second row show the imitation network makes this steering decision mainly following the curve of the double yellow line. However, when deliberate attack patterns are painted on the road as shown in Fig. \ref{fig:result03}(b), the imitation network fails to perceive the painted lines which could be easily ignored by a human; instead, the network regards the lines as physical barriers and guides the vehicle to steer left (steering -0.18) to avoid a collision, leading to a catastrophe. The reconstructed image below confirms that the most outstanding features are the painted adversaries instead of the central double yellow lines. We speculate that the vehicle recognizes the adversaries as the road curb. And Fig. \ref{fig:result03}(c) confirms our speculations. In this case, the vehicle is turning left and the corresponding reconstructed image shows the curb would contribute the strongest activations in the network which will make the steering angle a negative value (steering -0.24) to turn left. The similarity of the reconstructed inputs between cases (b) and (c) suggests that the painted attacks are misrecognized as a curb which leads to an unwise driving decision. To summarize, the deliberate adversaries that mimic important road features are very likely to be able to successfully attack the imitation learning algorithm. This also emphasizes the importance of taking more diverse training samples into consideration when designing autonomous driving techniques. Note that since the imitation learning network makes driving decisions solely based on current camera input, using one frame per case for visualization is enough to unravel the root causes of an attack's success.

\section{Conclusion}
In this paper, we develop a versatile modeling framework and simulation infrastructure to study adversarial examples on e2e autonomous driving models.
Our model and simulation framework can be applied beyond the scope of this paper, providing useful tools for future research to expose latent flaws in current models with the ultimate goal of improving them.
Through comprehensive experiment results, we demonstrate that simple physical adversarial examples that are easily realizable, such as mono-colored single-line and double-line patterns, not only exist, but can be quite effective under certain driving scenarios, even for models that perform robustly  without any attacks.
Our analysis using the DeConvNet method offers critical insights to further explore attack generation and defense mechanisms.
We plan to open-source our pattern generator upon the publication of this paper.

\bibliographystyle{ieeetr}
\bibliography{bibliography}
\end{document}